\documentclass[10pt,twocolumn,letterpaper]{article}

\usepackage[pagenumbers]{cvpr} 
\usepackage{multirow}
\usepackage{comment}
\usepackage{caption}
\usepackage{cuted}
\usepackage[table]{xcolor}
\usepackage{tabularx}
\usepackage{algorithm}
\usepackage{algorithmic}

\definecolor{cvprblue}{rgb}{0.21,0.49,0.74}
\usepackage[pagebackref,breaklinks,colorlinks,allcolors=cvprblue]{hyperref}

\def\paperID{9655} 
\def\confName{CVPR}
\def\confYear{2026}

\title{MG-Nav: Dual-Scale Visual Navigation via Sparse Spatial Memory}

\author{
Bo Wang$^{1,*}$ \quad
Jiehong Lin$^{1,*}$ \quad
Chenzhi Liu$^{1}$ \quad
Xinting Hu$^{1}$ \quad
Yifei Yu$^{1}$ \quad
Tianjia Liu$^{1}$ \\
Zhongrui Wang$^{2,\dagger}$ \quad
Xiaojuan Qi$^{1,\dagger}$ \\
{\tt\small wangzr@sustech.edu.cn} \quad {\tt\small xjqi@eee.hku.hk} \\
$^{1}$The University of Hong Kong \quad
$^{2}$Southern University of Science and Technology \\
$^*$Equal contribution \quad
$^\dagger$Corresponding authors
}

\begin{document}
\maketitle


\begin{strip}
\centering
\includegraphics[width=0.9\linewidth]{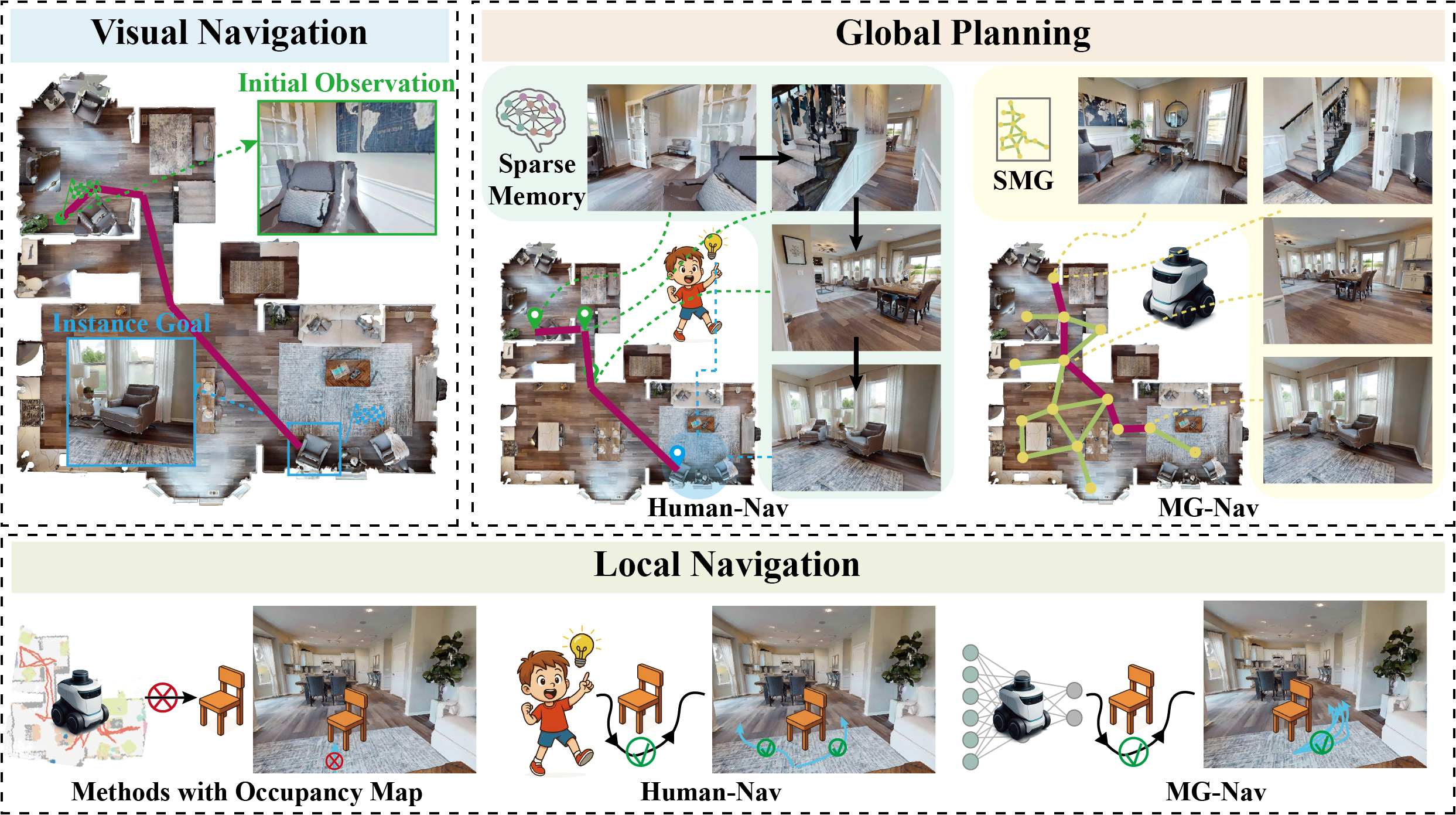}
\captionof{figure}{Overview of the proposed \textbf{MG-Nav}, a dual-scale framework that unifies global planning and local control for zero-shot visual navigation.
(a) \textbf{Global planning:} MG-Nav plans over a \textit{Sparse Spatial Memory Graph} (SMG), a compact region-centric memory that mirrors human navigation by providing node-level guidance without requiring dense 3D reconstruction. 
(b) \textbf{Local navigation:} MG-Nav employs navigation foundation policies enhanced with VGGT geometric features to improve goal recognition and enable obstacle-aware control.
MG-Nav operates global planning and local navigation at different frequencies and uses periodic re-localization to correct errors, which effectively handles dynamic changes and avoids collisions compared to methods that rely on dense 3D reconstruction.
}
\label{img:teaser}
\end{strip}

\begin{abstract}

We present \textbf{MG-Nav} (Memory-Guided Navigation), a dual-scale framework for zero-shot visual navigation that unifies global memory-guided planning with local geometry-enhanced control. At its core is the \textit{Sparse Spatial Memory Graph} (SMG), a compact, region-centric memory where each node aggregates multi-view keyframe and object semantics, capturing both appearance and spatial structure while preserving viewpoint diversity. At the global level, the agent is localized on SMG and a goal-conditioned node path is planned via an image-to-instance hybrid retrieval, producing a sequence of reachable waypoints for long-horizon guidance. At the local level, a navigation foundation policy executes these waypoints in point-goal mode with obstacle-aware control, and switches to image-goal mode when navigating from the final node towards the visual target. To further enhance viewpoint alignment and goal recognition, we introduce \textbf{VGGT-adapter}, a lightweight geometric module built on the pre-trained VGGT model, which aligns observation and goal features in a shared 3D-aware space. 
MG-Nav operates global planning and local control at different frequencies, using periodic re-localization to correct errors.
Experiments on HM3D Instance-Image-Goal and MP3D Image-Goal benchmarks demonstrate that MG-Nav achieves state-of-the-art zero-shot performance and remains robust under dynamic rearrangements and unseen scene conditions.

\end{abstract}

\section{Introduction}

Navigation is a fundamental capability for embodied agents to interact intelligently with their surroundings~\cite{mirowski2016learning,wong2025survey}. From domestic robots and delivery drones to AR/VR telepresence systems, the ability to reach a visual target in previously unseen environments, without explicit maps or dense supervision, remains a key milestone toward general embodied intelligence. Within this context, \textbf{visual navigation}, where the goal is specified by an image depicting either a target object instance~\cite{IEVE,ModIIN,Unigoal} or a scene~\cite{imagenav1,imagenav2,imagena3}, is particularly valuable yet challenging. The agent must infer the underlying 3D structure from a single goal image, reason over novel spatial layouts, and accurately reach the corresponding viewpoint or object while avoiding collisions and adapting to dynamic scene changes.

Existing frameworks for visual navigation can be broadly categorized into three families, each facing inherent limitations in unseen scenarios.
\textit{(i) Foundation policy models}~\cite{Nomad,NavDP} leverage large-scale trajectory pretraining to achieve strong generalization, yet struggle when the goal is invisible, often degenerating into unguided exploration with weak long-range reasoning~\cite{ViNT,schmittle2025long}.
\textit{(ii) Reinforcement Learning (RL) methods}~\cite{FGPrompt,REGNav,RSRNav} excel in fine-grained control within known environments but demand massive interaction data and fail to generalize under domain shifts.
\textit{(iii) Memory-based zero-shot approaches} construct persistent global maps~\cite{Goat,IEVE,bscnav} or scene graphs~\cite{Unigoal} to enable long-horizon planning without retraining. However, these methods typically rely on dense RGB-D reconstructions, which are expensive to build and brittle to even mild rearrangements or dynamic changes after memory construction.


In light of these limitations, it is instructive to revisit how humans navigate, which offers a striking contrast and source of inspiration. Unlike artificial agents that depend on dense maps or precise metric reconstructions, humans navigate effectively through \textit{sparse visual memories}, a limited set of distinctive snapshots that anchor our sense of place and orientation~\cite{spatial_memory1,spatial_memory2,spatial_memory3}. Such memories provide coarse yet reliable global guidance, enabling us to recall spatial relations and approximate locations even after long intervals or environmental changes.
Concurrently, humans engage in \textit{continuous local replanning}, dynamically adapting motion to avoid obstacles, resolve occlusions, and respond to unforeseen scene variations~\cite{spatial_memory1,spatial_memory2}.
This hierarchical strategy, \ie, \textit{global guidance from sparse memory coupled with local reactive contron}, forms the foundation of robust and adaptive navigation in complex, dynamic environments.

Inspired by this, we propose \textbf{MG-Nav} (\textit{Memory-Guided Navigation}), a dual-scale framework that integrates global memory-guided planning with local geometry-enhanced control. Fig. \ref{img:teaser} gives an overview of MG-Nav. At its core lies the \textbf{Sparse Spatial Memory Graph (SMG)}, a compact, region-centric representation of the explored environment. Each node in the SMG corresponds to a spatial region and aggregates a small set of multi-view keyframes together with instance-level semantics, while edges encode navigable connectivity between regions. This sparse abstraction mirrors human navigation practice, capturing distinctive visual memories that preserve viewpoint diversity and semantic consistency without relying on dense 3D reconstruction. 

At the \textit{global level}, MG-Nav performs high-level planning over SMG. Given a visual goal, the agent is first localized on SMG through an \textit{image-to-instance hybrid retrieval mechanism} that jointly matches both the current observation and the goal image to their corresponding nodes. Based on the matched nodes, a goal-conditioned node path is planned along the edges of SMG, yielding a sequence of reachable waypoints that provide global guidance and decompose the long-horizon navigation into node-to-node traversals. 

At the \textit{local level}, we employ a zero-shot foundation policy, pre-trained on large-scale data, which demonstrates strong obstacle avoidance and robustness to dynamic scene changes, to execute motion between adjacent SMG nodes. To further enhance geometric reasoning and visual goal alignment, we introduce \textbf{VGGT-adapter}, which incorporates geometry-aware features from the foundational Visual Geometry Group Transformer (VGGT) model \cite{vggt} into the policy. This integration preserves spatial consistency under rapid viewpoint changes and improves precise alignment with the visual goal, particularly during the final approach. 

Global planning and local control operate \textit{at different frequencies}, with periodic re-localization and re-planning, to absorb execution noise and recover from errors. By seamlessly combining global node-level guidance with local, goal-directed control, MG-Nav achieves robust zero-shot visual navigation in complex, dynamic environments.


Extensive experiments show that MG-Nav attains state-of-the-art results on both HM3D Instance-Image-Goal and MP3D Image-Goal, reaching 78.5/59.3 and 83.8/57.1 (SR/SPL).
Ablation studies verify the effectiveness of the different components of MG-Nav.
Moreover, in dynamic environments with newly inserted obstacles, MG-Nav remains stable with only minor performance drops, demonstrating strong robustness.




Our main contributions are summarized as follows:
\begin{itemize}
\item \textbf{Dual-scale memory-guided navigation.} We present MG-Nav, a unified architecture that integrates high-level semantic–topological planning with low-level geometry-enhanced control, enabling robust zero-shot navigation.
\item \textbf{Sparse Spatial Memory Graph (SMG).} A region-centric, multi-view memory graph representation that encodes both keyframe and object semantics, supporting efficient hybrid retrieval and node-level global planning.
\item \textbf{Geometric enhancement for local navigation.} We introduce a lightweight VGGT-adapter that aligns observation and goal embeddings in a shared 3D-aware space using VGGT features, substantially improving viewpoint robustness and goal matching precision.
\item \textbf{Empirical results.} MG-Nav achieves the state-of-the-art zero-shot performance on HM3D/MP3D benchmarks, and maintains robustness under dynamic scene changes.
\end{itemize}

\label{sec:intro}

\section{Related Work}

\noindent \textbf{Memory-Free Foundation Policy Methods.}
Foundation policies (e.g., GNM, ViNT, NoMaD, NavDP~\cite{Gnm,ViNT,Nomad,NavDP}) provide reliable short-horizon, zero-shot control but remain memory-free, relying on reactive visual similarity without global state and leading to failures when the goal is out of view and weak long-range reasoning.

\noindent \textbf{Memory-Free Reinforcement Learning Methods.}
End-to-end RL approaches \cite{FGPrompt, REGNav, RSRNav, RL_based, OVRL} target fine-grained control and incorporate LLMs for high-level command interpretation (e.g., CompassNav~\cite{li2025compassnav}).  
Yet they lack explicit global memory and often generalize poorly to unseen layouts and object configurations, struggling to maintain instance–image goal consistency over long horizons.


\noindent \textbf{Memory-Based Methods.}
For long-range planning, memory-based methods build global scene representations. Metric or dense maps (e.g., BSC-Nav, GOAT, IEVE, MOD-IIN~\cite{bscnav,Goat,IEVE,ModIIN}) and neural volumetric models (e.g., GaussNav, GSplatVNM~\cite{gaussnav,gsplatvnm}) offer accurate pose fidelity but are heavy and brittle under rearrangements. Sparse graphs are more efficient, yet object-centric graphs (e.g., UniGoal~\cite{Unigoal}) lose local context, while topology-heavy systems (e.g., Astra, Mobility-VLN~\cite{Astra,Mobility}) add complexity by intertwining graph reasoning with LLM-based planning. We address these trade-offs with a \emph{sparse, region-centric} Scene Memory Graph (SMG) that preserves local context for hybrid retrieval and a dual-scale design combining SMG-based global planning with robust local control.

\begin{figure*}[t]
\centering
\includegraphics[width=0.97\linewidth]{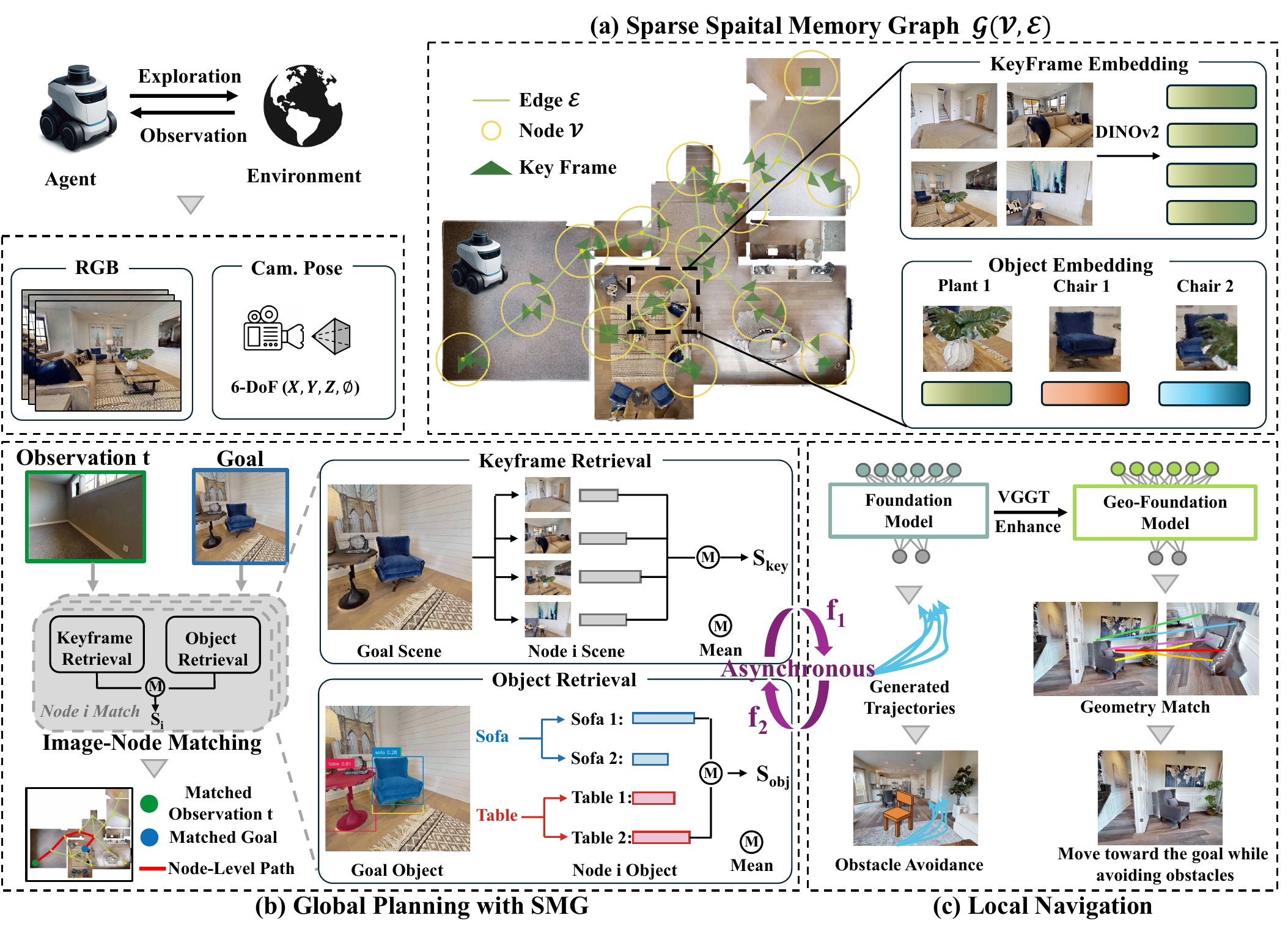}
\vspace{-0.4cm}
\caption{Illustration of the navigation process of \textbf{MG-Nav}, a dual-scale framework combining global planning with local execution. 
    \textbf{(a) Sparse Spatial Memory Graph (SMG)} serves as a compact, region-centric memory; each node aggregates multi-view keyframes and object semantics, while edges encode navigable connectivity. 
    \textbf{(b)  Global Planning with SMG}: Both the agent and the goal are localized on the SMG via an image-to-instance hybrid node retrieval. A goal-conditioned path from the current observation at time t to the goal is then planned along the graph edges to provide global guidance.
    \textbf{(c) Local Navigation via Geometry-Enhanced Policy}: a navigation foundation policy, geometrically enhanced with the \textbf{VGGT-adapter}, moves the agent between adjacent nodes while maintaining obstacle avoidance and accurate visual goal alignment. By running global planning (f$_1$) and local navigation (f$_2$) at different frequencies with periodic re-localization, MG-Nav achieves robust zero-shot navigation in dynamic, unseen environments.
}
\label{fig2:method}
\end{figure*}

\section{Memory-Guided Navigation}

\subsection{Overview}

Visual navigation aims to predict a sequence of $M$ actions $\mathcal{A} = (\mathbf{a}_1, \dots, \mathbf{a}_M)$ that guide an agent to a visual goal $\mathbf{I}_{goal}$ depicting the target object or scene, where each action is defined by planar displacement and orientation. To address this task, we propose \textbf{MG-Nav} (\textbf{Memory-Guided Navigation}), a dual-scale framework that integrates global planning with local execution. 

An overview of the MG-Nav workflow is shown in Fig. \ref{fig2:method}. The core idea is that a small number of distinctive visual memories provide coarse global guidance, while local motion is continuously adapted to avoid obstacles and dynamic changes. To embody this idea, we introduce \textbf{Sparse Spatial Memory Graph} (SMG) as an abstract scene representation (Sec. \ref{sec:smg}). Formally, SMG is defined as $\mathcal{G} = (\mathcal{V}, \mathcal{E})$, where nodes $\mathcal{V}$ function as sparse, memorable spatial anchors, and edges $\mathcal{E}$ encode their navigable connectivity. Upon $\mathcal{G}$, MG-Nav performs dual-scale navigation: (i) \underline{\textit{global planning}}, which retrieves the memory node $\mathbf{v}_{goal} \in \mathcal{V}$ best aligned with the goal image $\mathbf{I}_{goal}$ and plans a node-level path $(\mathbf{v}_1, \dots, \mathbf{v}_{K-1}, \mathbf{v}_K)$ along $\mathcal{E}$ (Sec. \ref{sec:globalplan}), where $\mathbf{v}_K$ corresponds to the exact visual goal $\mathbf{I}_{goal}$ with $\mathbf{v}_{K-1} = \mathbf{v}_{goal}$; and (ii) \underline{\textit{local navigation}}, where a geometry-enhanced diffusion policy navigates between consecutive nodes to constitute the whole action sequence $\mathcal{A} = (\mathcal{A}_1, \dots, \mathcal{A}_{K-1}, \mathcal{A}_K)$ (Sec. \ref{sec:localnav}). 
By operating global planning and local navigation at different frequencies, with periodic re-localization and re-planning, MG-Nav mitigates execution noise and drift, and thus enables robust zero-shot navigation in dynamic and previously unseen environments (Sec.~\ref{sec:loop}).

\begin{figure}[t]
\centering
\includegraphics[width=0.96\linewidth]{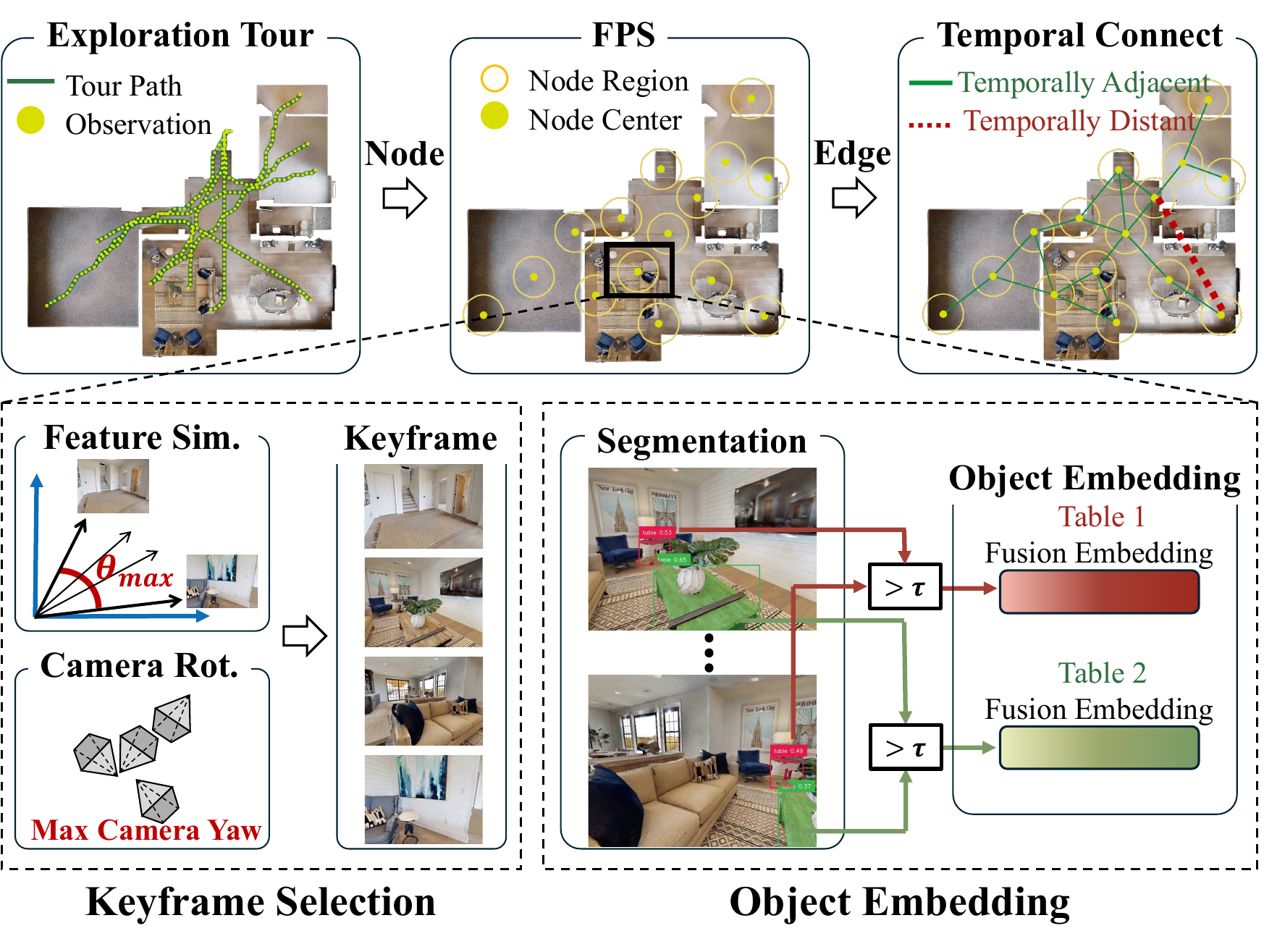}
\vspace{-0.4cm}
\caption{Illustration of the construction of \textbf{Sparse Spatial Memory Graph}. Each node in SMG represents a spatial region, aggregating a small set of both multi-view keyframe and object embeddings, while edges between nodes encode navigable connectivity.
}
\label{fig3:smgconstruct}
\end{figure}

\subsection{Sparse Spatial Memory Graph}
\label{sec:smg}

Sparse Spatial Memory Graph (SMG) $\mathcal{G} = (\mathcal{V}, \mathcal{E})$ forms the core of MG-Nav to enable global planning. The construction process of $\mathcal{G}$ is illustrated in Fig. \ref{fig3:smgconstruct}. 

More specifically, for each indoor scene, we first follow standard practice to collect posed tour demonstrations \cite{bscnav,gaussnav}. We then apply Farthest-Point Sampling (FPS) to the extrinsic camera poses of the demonstration frames to obtain sparse but spatially representative locations. Around each sampled location, we define a region of radius $r$ as a spatial node $\mathbf{v} \in \mathcal{V}$, while the edge set $\mathcal{E}$ is derived from temporal adjacency during the tours, which naturally ensures that connected nodes correspond to mutually reachable regions in the physical environment.

We further represent each spatial node as a \textit{structural embedding} $\mathbf{v} = (\mathbf{P}, \mathbf{F}, \mathbf{O})$, where $\mathbf{P} \in \mathbb{R}^3$ denotes the 3D center of the region, $\mathbf{F} \in \mathbb{R}^{N_f \times C}$ contains $N_f$ keyframe embeddings, and $\mathbf{O} \in \mathbb{R}^{N_o \times C}$ contains $N_o$ representative object embeddings, with $C$ as the feature dimension. Concretely, $\mathbf{P}$ is set as the sampled camera location of $\mathbf{v}$, while $\mathbf{F}$ and $\mathbf{O}$ are obtained as follows:

\textbf{(i) Keyframe embeddings - }We extract the \texttt{CLS} token from DINOv2 \cite{oquab2023dinov2} for demonstration frames within the local region of $\mathbf{v}$. To ensure diversity, we first select $N_f'$ frames with the lowest feature similarity and then pick the final $N_f$ frames with the largest camera rotation variance among them. The embeddings of these selected frames form $\mathbf{F}$, providing rich regional context.

\textbf{(ii) Object embeddings - }We apply the open-vocabulary segmentation model Grounded-SAM \cite{Groundedsam} to all frames within the region of $\mathbf{v}$ to extract object instances. For each segmented object, we use DINOv2 to obtain embeddings from the \texttt{CLS} token and the average patch features, which are combined via a weighted summation to form robust object embeddings. Objects of the same category whose embedding cosine similarity exceeds a threshold $\tau$ are further considered the same instance, and their embeddings are averaged to produce a unique feature in $\mathbf{O}$.

SMG provides a sparse yet spatially representative structure that integrates information from both multi-view keyframes and object instances, thus preserving distinctive visual memory anchors to simplify the planning space.

\subsection{Global Planning with SMG}
\label{sec:globalplan}

As shown in Fig. \ref{fig2:method}, global planning on SMG $\mathcal{G} = (\mathcal{V}, \mathcal{E})$ consists of two steps, including (i) matching both the current observation $\mathbf{I}_{obs}$ and the visual goal $\mathbf{I}_{goal}$ to nodes in $\mathcal{V}$, and (ii) planning a node-level trajectory along SMG edges that provides high-level guidance for local navigation.

\subsubsection{Image-Node Matching}
Given an input image ($\mathbf{I}_{obs}$ or $\mathbf{I}_{goal}$), we aim to retrieve the node in $\mathcal{V}$ most semantically and visually aligned with it, using a \textit{hybrid retrieval} strategy combining global scene similarity and object-level semantics:


\textbf{(i) Keyframe retrieval -} For each node $\mathbf{v} \in \mathcal{V}$, we compute the cosine similarity between the DINOv2 \texttt{CLS} embedding of the input image and each of the $N_f$ keyframe embeddings $\mathbf{F}$, taking the maximum similarity as the keyframe score of the node. The top-$N_o$ nodes with the highest keyframe scores are retained as candidates for subsequent object-level verification.

\textbf{(ii) Object retrieval -} Following the procedure in Sec. \ref{sec:smg}, we apply Grounded-SAM for open-vocabulary segmentation and DINOv2 for feature extraction to obtain object embeddings of the image. For each object embedding in the image, we compute its cosine similarity with object embeddings of the same category in each node candidate, taking the maximum score; if a node does not contain that category, the score is set to 0. For each node, we average the scores across all objects to obtain its object similarity score.

Finally, for the top-$N_o$ node candidates, we average their keyframe and object similarity scores, and select the node with the highest combined score as the retrieval result.

\subsubsection{Node-Level Path Planning}

Based on the hybrid retrieval results, we obtain the matched nodes $\mathbf{v}_{obs}$ and $\mathbf{v}_{goal}$ corresponding to $\mathbf{I}_{obs}$ and $\mathbf{I}_{goal}$. We then perform high-level planning on SMG $\mathcal{G}$ by applying A* search \cite{A_star} over the edge set $\mathcal{E}$. Since each edge denotes a physically feasible transition observed in demonstrations, the resulting path provides a reliable structural prior for navigation. The planned node-level trajectory is denoted as $(\mathbf{v}_1, \dots, \mathbf{v}_{K-1}, \mathbf{v}_K)$, where $\mathbf{v}_1 = \mathbf{v}_{obs}$ and $\mathbf{v}_{K-1} = \mathbf{v}_{goal}$; we additionally append $\mathbf{v}_K$ to represent $\mathbf{I}_{goal}$, ensuring that the node sequence explicitly terminates at the goal state.

\subsection{Local Navigation via Geometry-Enhanced Policy}
\label{sec:localnav}

The node-level path planned on SMG decomposes navigation into a sequence of intermediate sub-goals, thereby alleviating the difficulty of fine-grained motion control toward the final visual goal. Building upon this, we further address the zero-shot local navigation between adjacent nodes using general navigation foundation policies, and introduce \textbf{VGGT-adapter} for geometric enhancement to improve visual goal matching capacity of policies.

\subsubsection{Geometry-Enhanced Policy with VGGT-adapter}

Navigation foundation policies are typically formulated as $\mathcal{A} = \pi(\mathbf{I}_{obs}, \mathbf{P}_{goal}, \mathbf{I}_{goal})$, where the goal is specified either as a point $\mathbf{P}_{goal}$ or an image $\mathbf{I}_{goal}$. Although these policies are trained on large-scale trajectories and demonstrate strong short-horizon navigation capabilities with obstacle avoidance and robustness to dynamic scene changes, they still struggle to precisely localize and pursue visual goals, especially under significant viewpoint variations.

To mitigate this limitation, we introduce the \textbf{VGGT-adapter}, which integrates geometry-aware features derived from the pre-trained Visual Geometry Group Transformer (VGGT) model \cite{vggt} into the policy. VGGT inherently models multi-view 3D structure and pixel correspondences, making it suitable for enhancing spatial reasoning and improving visual goal matching during navigation.

More specifically, we use VGGT to extract geometry-aware feature maps from both the current observation $\mathbf{I}_{obs}$ and the visual goal $\mathbf{I}_{goal}$. After applying spatial average pooling, the resulting feature maps are flattened into token sequences $\mathbf{F}_{obs}^{G}$ and $\mathbf{F}_{goal}^{G}$. The two token sequences are then concatenated and passed through the VGGT-adapter, implemented as a lightweight multi-layer perceptron (MLP), which projects the geometry tokens into the policy embedding space as $\mathbf{F}_{adpt}^{G} = \text{MLP}([\mathbf{F}_{obs}^{G} \, | \, \mathbf{F}_{goal}^{G}])$. Finally, $\mathbf{F}_{adpt}^{G}$ is concatenated with the original visual condition tokens of the navigation policy, thereby enriching it with 3D geometric cues and inter-frame correspondence awareness to strengthen visual goal matching.

\subsubsection{Node-based Local Navigation}

With the node-level path $\{\mathbf{v}_1,\dots,\mathbf{v}_K\}$ obtained from SMG, our geometric-enhanced policy $\pi$ is employed to perform local navigation in a node-to-node manner to progressively approach the final visual goal while avoiding obstacles. For each step $k$, the agent navigates from the current node $\mathbf{v}_{k-1}$ to the next node $\mathbf{v}_{k}$ by predicting a chunk of executable actions:
\begin{equation}
\mathcal{A}_k =
\begin{cases}
\pi(\mathbf{I}_k, \mathbf{P}_k, -) & k \ne K \\
\pi(\mathbf{I}_k, -, \mathbf{I}_{goal}) & k = K
\end{cases},
\end{equation}
where $\mathbf{I}_k$ denotes the current observation and $\mathbf{P}_k$ is the 3D location of node $\mathbf{v}_k$. 
Accordingly, the policy operates in \textit{point-goal mode} when moving between intermediate nodes, and switches to \textit{image-goal mode} only at the final stage to precisely align with the visual goal $\mathbf{I}_{goal}$, befitting from the geometric enhancement of VGGT-adapter. The overall navigation action sequence is thus given as $\mathcal{A} = (\mathcal{A}_1, \dots, \mathcal{A}_{K-1}, \mathcal{A}_K)$. 

\subsection{Dual-scale Planning Loop}
\label{sec:loop}

MG-Nav coordinates planning and control at two frequencies: a slow global loop (every $T_g$ steps) and a fast local loop (every $T_\ell \ll T_g$). The global loop re-localizes the agent on SMG via hybrid retrieval and updates the node-level path (Sec. \ref{sec:globalplan}) whenever confidence drops or edges become blocked, yielding an updated node sequence. The local loop continuously executes short-horizon actions toward the current node while performing obstacle-aware control (Sec. \ref{sec:localnav}). At runtime, MG-Nav alternates between two tightly coupled processes:
\begin{itemize}
    \item \textbf{Local navigation} (Sec. \ref{sec:localnav}): The geometry-enhanced policy executes $\mathcal{A}_k$ to reach the current node $\mathbf{v}_k$, updating visual feedback at each step.
    \item \textbf{Periodic global re-localization and planning} (Sec. \ref{sec:globalplan}): Every $T_g$ local steps—or when the visual-confidence score drops—the agent is re-localized on SMG and A* is re-invoked to refine the remaining path.
\end{itemize}
This asynchronous, dual-frequency scheme mitigates drift, adapts to dynamic scene changes, and preserves long-horizon goal intent with minimal computational overhead.

\section{Experiments}

\subsection{Experimental Setup}
\noindent \textbf{Benchmarks:} Our method is tested on two 3D scene datasets, Habitat-Matterport 3D (HM3D)~\cite{ramakrishnan2021hm3d} for instance image-goal navigation (InstanceImageNav) and Matterport3D (MP3D)~\cite{Matterport3D} for image-goal navigation (ImageNav). Both tasks require the agent to navigate toward the target depicted in a given goal image, but ImageNav targets an entire scene or viewpoint, while InstanceImageNav specifies a particular object instance. For HM3D, we conduct experiments on 1000 episodes of 36 validation scenes following the Instance ImageNav-v3. For MP3D ImageNav, we conduct experiments on 1014 episodes of 5 testing scenes.

\noindent \textbf{Evaluation Metrics:} We evaluate with Success Rate (SR) to measure the proportion of successful episodes, and Success Rate weighted by Path Length (SPL) \cite{Unigoal, ModIIN} to assess navigation efficiency by accounting for path optimality.


\noindent \textbf{Implementation Details:} Our agent is deployed in Habitat Simulator~\cite{puig2023habitat3}. We employ DINOv2-ViT-L/14~\cite{oquab2023dinov2} for visual encoding, Grounded-SAM-2~\cite{Groundedsam} for semantic segmentation, and VGGT~\cite{vggt} for geometric embedding extraction. We use NavDP \cite{NavDP} as the foundation policy for MG-Nav's local navigation module. The success distance threshold is $1.0$ m and the maximum episode length is $500$ steps. Hyperparameters are detailed in appendix.


\begin{table}[t]
\centering
\caption{\textbf{InstanceImageNav on HM3D.}
Foundation models are shown on top, RL-based methods are shown on middle, and Memory-based methods are shown on bottom. }
\label{tab:insinav_hm3d}
\setlength{\tabcolsep}{2pt}
\begin{tabular}{lcccc}
\toprule
\textbf{Method} & \textbf{Memory} & \textbf{SR $\uparrow$} & \textbf{SPL $\uparrow$} \\
\midrule
ViNT\textsuperscript{*}~\cite{ViNT}    & --   & 7.7 & 6.7    \\
GNM\textsuperscript{*}~\cite{Gnm}    & --   & 11.4 & 5.6    \\
NoMAD\textsuperscript{*}~\cite{Nomad}   & --   & 16.8 & 7.0    \\
NavDP\textsuperscript{*}~\cite{NavDP}   & --   & 24.7 & 12.6    \\
\midrule\midrule
RL Baseline$^{\dagger}$~\cite{RL_based}      & --    & 8.3  & 3.5 \\
FGPrompt$^{\dagger}$~\cite{FGPrompt}         & --    & 9.9  & 2.8 \\
OVRL-v2 IIN$^{\dagger}$~\cite{OVRL}      & --    & 24.8  & 11.8 \\
CompassNav~\cite{li2025compassnav}       & --    & 35.6  & 14.8  \\
\midrule\midrule
GOAT~\cite{Goat}              & Scene Map (RGBD) & 37.4  & 16.1  \\
MOD-IIN~\cite{ModIIN}          & Scene Map (RGBD)   & 56.1  & 23.3  \\
UniGoal~\cite{Unigoal}          & Object Graph (RGBD) & 60.2  & 23.7  \\
IEVE~\cite{IEVE}            & Scene Map (RGBD) & 70.2  & 25.2  \\
BSC-Nav~\cite{bscnav}         & Scene Map (RGBD)  & 71.4  & 57.2  \\
GaussNav~\cite{gaussnav}        & 3DGS Map (RGBD)    & 72.5  & 57.8  \\
\textbf{MG-Nav(Ours)} & \textbf{SMG (RGB)} & \textbf{78.5} & \textbf{59.3} \\
\bottomrule

\multicolumn{4}{l}{\footnotesize \textsuperscript{*} Result from our re-implementation following their official code.} \\
\multicolumn{4}{l}{\footnotesize $^{\dagger}$ Result from GaussNav.} \\
\end{tabular}

\end{table}

\begin{table}[t]
\centering
\caption{\textbf{ImageNav on MP3D.}
Foundation models are shown on top, RL-based methods are shown on middle, and Memory-based methods are shown on bottom.}
\label{tab:inav_mp3d}
\begin{tabular}{lcccc}
\toprule
\textbf{Method} & \textbf{Memory} & \textbf{SR $\uparrow$} & \textbf{SPL $\uparrow$} \\
\midrule
ViNT\textsuperscript{*}~\cite{ViNT}  & -- & 6.12 & 4.98 \\
GNM\textsuperscript{*}~\cite{Gnm}  & -- & 10.06 & 7.36 \\
NoMAD\textsuperscript{*}~\cite{Nomad}  & -- & 10.95 & 5.64 \\
NavDP\textsuperscript{*}~\cite{NavDP}  & -- & 15.49 & 7.96 \\
\midrule\midrule
RSFG\textsuperscript{*}~\cite{RSFG} & -- & 68.54 & 43.18 \\
FGPrompt-MF\textsuperscript{*}~\cite{FGPrompt}  & -- & 71.20 & 39.74 \\
REGNav\textsuperscript{*}~\cite{REGNav} & -- & 74.66 & 47.24 \\
FGPrompt-EF\textsuperscript{*}~\cite{FGPrompt}  & -- & 77.71 & 51.09 \\
\midrule\midrule
\textbf{MG-Nav(Ours)} & \textbf{SMG (RGB)} & \textbf{83.77} & \textbf{57.15} \\
\bottomrule
\multicolumn{4}{l}{\footnotesize \textsuperscript{{*}} Result from our re-implementation following their official code.}
\end{tabular}
\end{table}

\begin{figure*}[ht]
\centering
\includegraphics[width=0.97\linewidth]{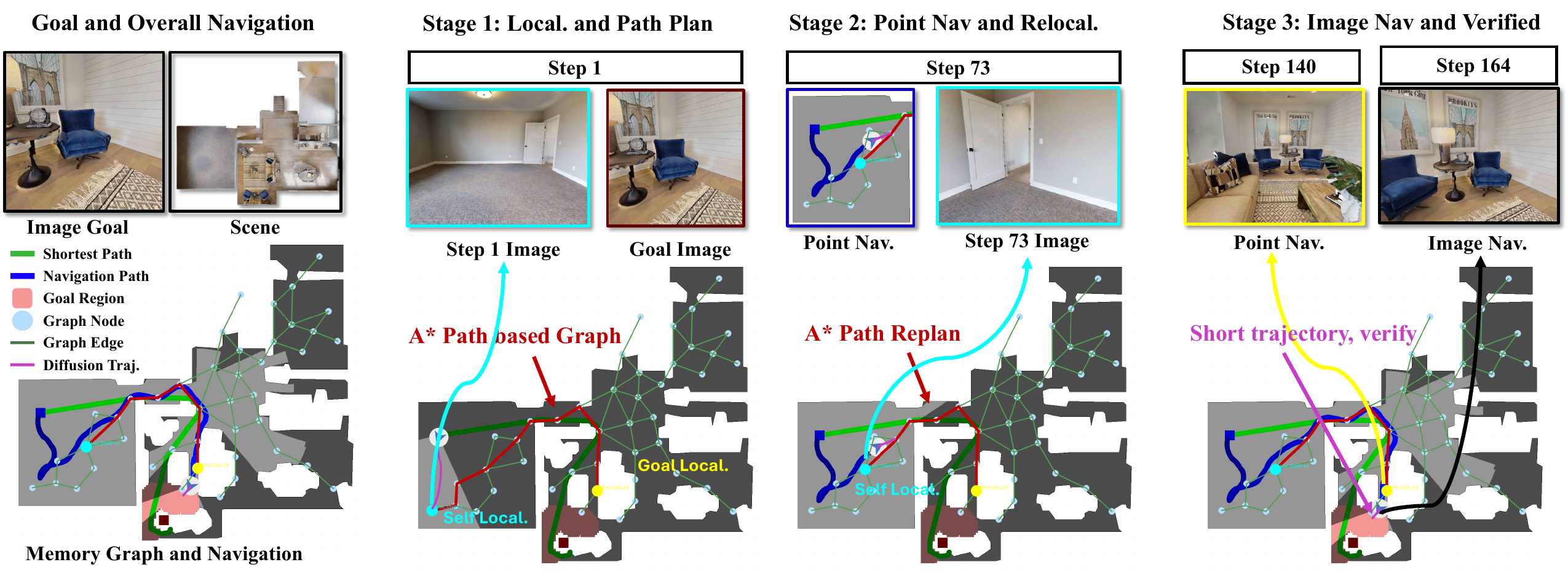}
\vspace{-0.3cm}
\caption{{Illustration of the decision process of MG-Nav.} Step 1 shows initial self- and goal-localization on SMG followed by global planning. 
Step 73 shows node-to-node navigation using point mode of NavDP, with periodic global re-localization. 
Step 140 shows the agent entering the matched goal-node region. 
Step 164 shows the policy switching to image mode and successfully verifying the target.} 
\label{Mg-Nav ins example}
\end{figure*}

\subsection{Main Results}
\paragraph{HM3D InstanceImageNav.}
Table~\ref{tab:insinav_hm3d} reports HM3D InstanceImageNav results.
Foundation policies (\eg, NavDP \cite{NavDP}, 24.7/12.6) and RL methods (\eg, CompassNav \cite{li2025compassnav}, 35.6/14.8) suffer from poor zero-shot generalization due to reactive local planning and weak global reasoning. While Memory-based methods (\eg, BSC-Nav \cite{bscnav}, 71.4/57.2; GaussNav \cite{gaussnav}, 72.5/57.8) achieve notable accuracy, they remain limited by maps that lack robust instance semantics and rely on unstable low-level planners. MG-Nav addresses these limitations by pairing a sparse, region-centric SMG with a foundation policy enhanced by a VGGT-based adapter for robust navigation and viewpoint alignment. This synthesis establishes a new state-of-the-art of SR 78.5 and yields shorter, more reliable paths (SPL 59.3). The successful example of the task executed by MG-Nav is shown in Fig.~\ref{Mg-Nav ins example}, and other examples are detailed in appendix.

\paragraph{MP3D ImageNav.}
Table~\ref{tab:inav_mp3d} reports overall MP3D ImageNav results.
MG-Nav attains 83.77/57.15 (SR/SPL), surpassing the strongest RL baseline FGPrompt-EF \cite{FGPrompt} 77.71/51.09 by 6.06/6.06. Foundation policies remain far lower, for example, NavDP \cite{NavDP} performs 15.49 and 7.96, where our method surpasses by 68.28/49.19. The huge improvement demonstrates that our dual-scale navigation system via SMG provide reliable global planning and robust view alignment, yielding state-of-the-art performance.


\begin{table*}[t]
\centering
\caption{\textbf{Ablation study of our method.}
We report SR/SPL while jointly exposing graph sparsity (node spacing $d$) and coverage radius $r$.
“Global” means keyframe retrieval, “Object” means object retrieval.} \vspace{-0.2cm}
\label{tab:ablation}
\setlength{\tabcolsep}{3pt}
\begin{tabular}{lccccccc}
\toprule
\textbf{Ablation} & \textbf{Variant} & \textbf{Graph} & \textbf{Retrieval} & \textbf{VGGT-adapter} & $\mathbf{d}$ (m) & $\mathbf{r}$ (m) & \textbf{SR $\uparrow$ / SPL $\uparrow$} \\
\midrule
\multirow{3}{*}{Component} 
    & Foundation Policy (NavDP) & $\times$ & --             & $\times$  & --   & --   & 24.70 \, / \, 12.60 \\
    & + SMG                     & \checkmark & Global+Object & $\times$  & 1.0  & 0.5  & 74.04 \, / \, 56.14 \\
    & + VGGT-adapter            & \checkmark & Global+Object & \checkmark& 1.0  & 0.5  & 78.50 \, / \, 59.27 \\
\midrule
\multirow{3}{*}{Retrieval} 
    & Instance Match only       & \checkmark & Object        & \checkmark& 1.0  & 0.5  & 72.20 \, / \, 52.79 \\
    & Global Match only         & \checkmark & Global        & \checkmark& 1.0  & 0.5  & 73.52 \, / \, 52.90 \\
    & Global + Instance Match   & \checkmark & Global+Object & \checkmark& 1.0  & 0.5  & 78.50 \, / \, 59.27 \\
\midrule
\multirow{3}{*}{Graph sparsity} 
    & Graph 1                   & \checkmark & Global+Object & \checkmark& 2.0  & 1.0  & 70.69 \, / \, 46.89 \\
    & Graph 2                   & \checkmark & Global+Object & \checkmark& 1.5  & 0.8  & 77.10 \, / \, 54.79 \\
    & Graph 3                   & \checkmark & Global+Object & \checkmark& 1.0  & 0.5  & 78.50 \, / \, 59.27 \\
\bottomrule
\end{tabular}
\end{table*}

\begin{figure*}[t]
\centering
\includegraphics[width=0.92\linewidth]{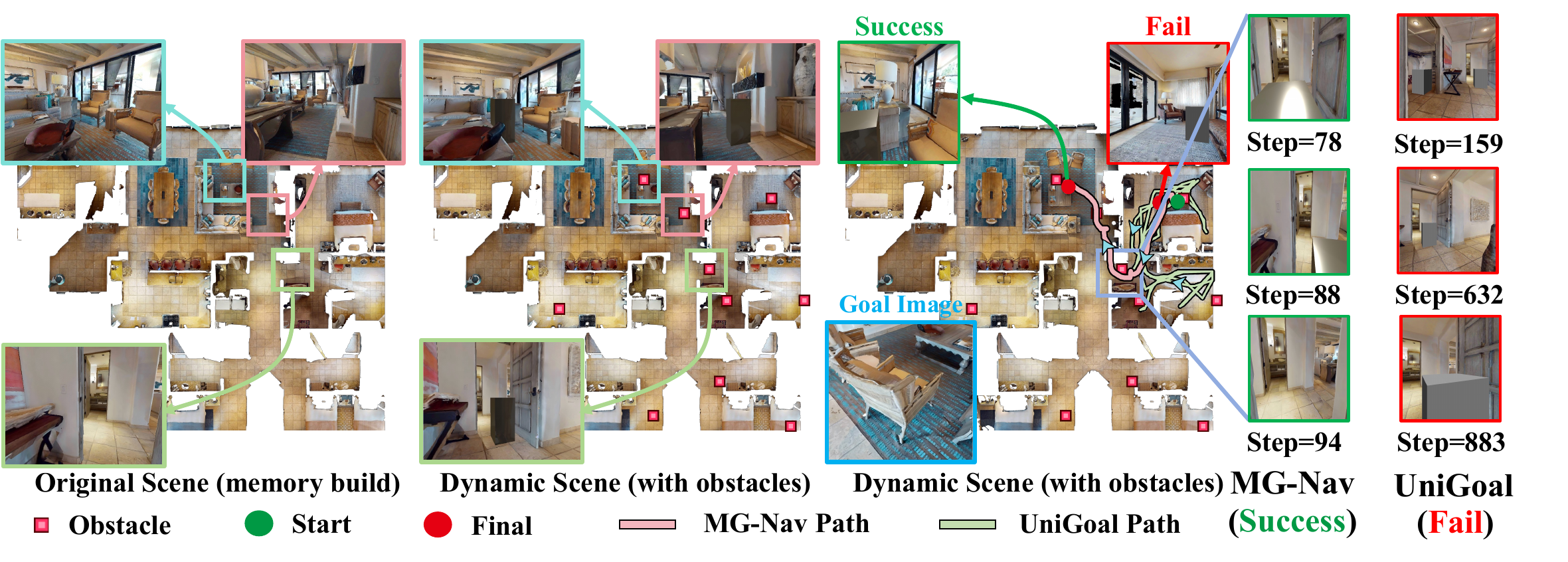}
\caption{Illustration of the robustness to dynamic scene changes of MG-Nav and UniGoal. 10 additional obstacles are added to scene \texttt{mv2HUxq3B53} (left) to model dynamic scenarios (middle). 
UniGoal becomes trapped near the inserted obstacles and keeps wandering in a local region until timeout (right, green path), whereas MG-Nav successfully avoids the newly added obstacles and reaches the goal (right, red path), demonstrating strong robustness to unmodeled scene rearrangements.
}
\label{fig5:obstacle}
\end{figure*}

\begin{table}[t]
\centering
\caption{\textbf{Robustness to dynamic scene changes.}
We evaluate each model on 100 InstanceImageNav episodes from HM3D under our dynamic-scene setting.
Results for UniGoal and BSC-Nav are obtained from our re-implementations based on their official code.}
\label{tab:robustness}
\setlength{\tabcolsep}{3pt}
\begin{tabular}{lcccccc}
\toprule
\textbf{Obs. Num} & \multicolumn{2}{c}{\textbf{0}} & \multicolumn{2}{c}{\textbf{5}} & \multicolumn{2}{c}{\textbf{10}} \\
\cmidrule(lr){2-3}\cmidrule(lr){4-5}\cmidrule(lr){6-7}
\textbf{Method} & \textbf{SR} & \textbf{SPL} & \textbf{SR} & \textbf{SPL} & \textbf{SR} & \textbf{SPL} \\
\midrule
BSC-Nav~\cite{bscnav}  & 25.49   & 19.91 & 8.64 & 4.63 & 7.84 & 4.94 \\
UniGoal~\cite{Unigoal}  & 56.43 & 20.44 & 52.94 & 19.68 & 44.21 & 17.17 \\
\textbf{Ours} & \textbf{73.53} & \textbf{56.28} & \textbf{72.55} & \textbf{52.20} & \textbf{68.63} & \textbf{50.15} \\
\bottomrule
\end{tabular}
\end{table}

\subsection{Ablation Study}
\noindent \textbf{Component ablation.}
Table~\ref{tab:ablation} reports a detailed ablation on model components. 
The foundation model achieves only 24.7/12.6 (SR/SPL), limited by its purely reactive control and lack of long-range planning.
Introducing SMG for global planning significantly boosts performance to 74.04/56.14, as it decomposes long-range navigation into reachable node-to-node subgoals, providing global guidance for the agent.
Further incorporating the VGGT-adapter improves results to 78.50/59.27, confirming that geometry-aware observation–goal alignment enhances robustness during viewpoint shifts and final approach stages.

\noindent \textbf{Retrieval strategy ablation.}
Our hybrid retrieval strategy, which combines keyframe (global) and object-level semantics, achieves the best performance (78.50/59.27 in SR/SPL). We perform an ablation to study the necessity of each component. Relying only on global keyframe retrieval degrades performance by a large margin (from 78.50/59.27 to 73.52/52.90), and using only object-level retrieval also results in a substantial drop (to 72.20/52.79). This confirms the need for a hybrid strategy. Global appearance enables coarse scene localization and long-range recall, while instance semantics enhance object-level discrimination and viewpoint invariance. Combining both yields more reliable and robust matching.

\noindent \textbf{Graph sparsity ablation.}
We analyze the impact of the SMG structure by adjusting node spacing ($d$) and coverage radius ($r$).
The configuration $(d{=}1.0\,\text{m}, r{=}0.5\,\text{m})$ achieves the best 78.50/59.27 (SR/SPL).
Making the graph moderately sparser to $(1.5\,\text{m}, 0.8\,\text{m})$ yields 77.10/54.79 ($-1.40$/$-4.48$).
Further sparsifying to $(2.0\,\text{m}, 1.0\,\text{m})$ degrades to 70.69/46.89 ($-7.81$/$-12.38$).
The sharper decline in SPL indicates that overly sparse graphs reduce viewpoint diversity and weaken topological connectivity, leading to longer local execution segments and detours.

\subsection{Robustness to Dynamic Scene Changes}
To assess robustness to unmodeled dynamic scene changes, we first construct the Memory Graph on the original HM3D scenes. We then insert different numbers (0/5/10) of random obstacles into the environment during the navigation phase to simulate a dynamic environment (Fig. \ref{fig5:obstacle}). These newly inserted objects serve as dynamic obstacles that the agent must robustly avoid to reach its goal. The performance of MG-Nav under these conditions is then compared with representative mapping-based methods (Table~\ref{tab:robustness}).
We observe that methods degrade sharply (\eg, BSC-Nav \cite{bscnav} SR 25.49 $\rightarrow$ 7.84, UniGoal \cite{Unigoal} SR 56.43 $\rightarrow$ 44.2), while MG-Nav shows only minor drops (SR 73.53 $\rightarrow$ 68.63 and SPL 56.28 $\rightarrow$ 50.15).
This resilience stems from a decoupled dual-scale design. The sparse SMG delivers robust region-level global planning, keeping the navigational goal stable, while the zero-shot local policy handles unmodeled obstacles and avoids them without relying on the global map, resulting in only minor drops in SR and SPL.

\section{Conclusion}

We introduced \textbf{MG-Nav}, a dual-scale navigation framework that combines global memory-guided planning with local geometry-enhanced control for zero-shot visual navigation. The core Sparse Spatial Memory Graph (SMG) provides a compact, region-centric representation, enabling long-horizon reasoning and robustness to moderate scene changes. Global planning retrieves goal-conditioned node paths, while a pre-trained navigation policy, augmented with the VGGT-adapter, executes node-to-node motion with enhanced visual and geometric alignment. Experiments show that MG-Nav achieves state-of-the-art zero-shot performance on challenging Image-Goal and Instance-Goal tasks, generalizing effectively to novel and dynamic environments. We believe that the dual-scale navigation with sparse spatial memory can inspire future research in scalable and robust embodied navigation, especially for agents operating in complex, unseen, or dynamic spaces.

\newpage
{
    \small
    \bibliographystyle{ieeenat_fullname}
    \bibliography{main}
}


\end{document}